\documentclass{article}
\usepackage{ijcai25}

\usepackage{times}
\usepackage{soul}
\usepackage{url}
\usepackage[hidelinks]{hyperref}
\usepackage[utf8]{inputenc}
\usepackage[small]{caption}
\usepackage{graphicx}
\usepackage{amsmath}
\usepackage{amsthm}
\usepackage{booktabs}
\usepackage{algorithm}
\usepackage{algorithmic}
\usepackage[switch]{lineno}

\usepackage{booktabs} 
\usepackage{multirow} 
\usepackage{adjustbox} 
\usepackage{amsmath} 
\usepackage{float}
\usepackage{booktabs} 
\usepackage{amssymb} 
\usepackage{natbib}

\title{SmartSpatial: Enhancing 3D Spatial Awareness in Stable Diffusion\\ with a Novel Evaluation Framework}

\author{
Mao Xun Huang$^1$\and
Brian J Chan$^2$\And
Hen-Hsen Huang$^3$\\
\affiliations
$^1$Department of Management Information Systems, National Chengchi University, Taipei, Taiwan\\
$^2$Department of Computer Science, National Chengchi University, Taipei, Taiwan\\
$^3$Institute of Information Science, Academia Sinica, Taipei, Taiwan\\
\emails
\{110306019,110703065\}@nccu.edu.tw,
hhhuang@iis.sinica.edu.tw
}

\let\oldtwocolumn\twocolumn
\renewcommand\twocolumn[1][]{%
    \oldtwocolumn[{#1}{
    \begin{center}
           \includegraphics[width=\linewidth]{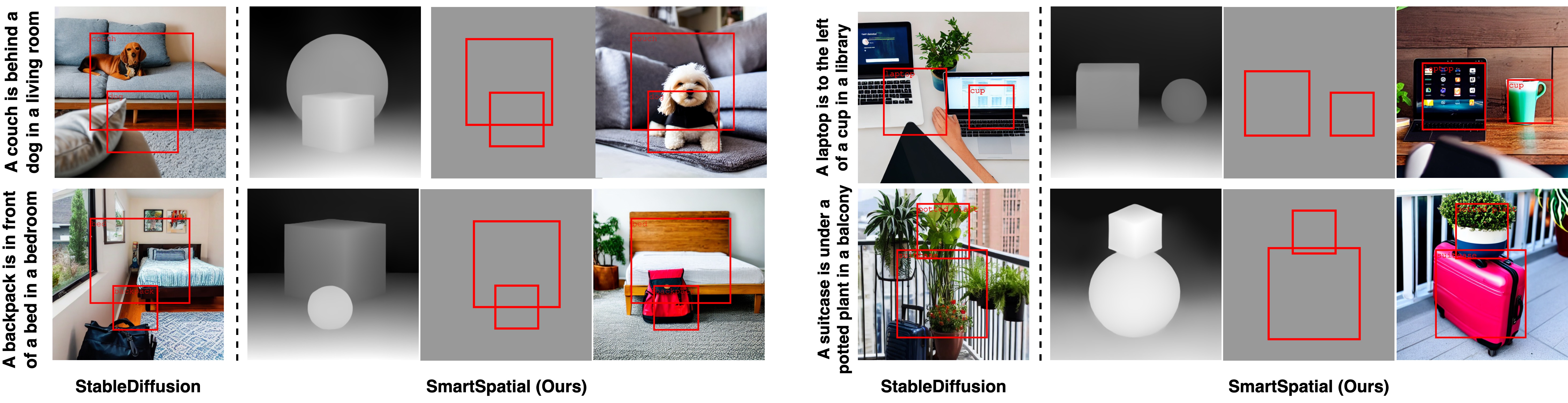}
           \captionof{figure}{Example images generated using Stable Diffusion (left) and SmartSpatial (right). With the provided depth map and layout control, SmartSpatial achieves superior spatial arrangement without requiring additional training or fine-tuning.}
           \label{fig:teaser}
        \end{center}
    }]
}
\begin{document}
\maketitle
\begin{abstract}
Stable Diffusion models have made remarkable strides in generating photorealistic images from text prompts but often falter when tasked with accurately representing complex spatial arrangements, particularly involving intricate 3D relationships. 
To address this limitation, we introduce SmartSpatial, an innovative approach that not only enhances the spatial arrangement capabilities of Stable Diffusion but also fosters AI-assisted creative workflows through 3D-aware conditioning and attention-guided mechanisms. 
SmartSpatial incorporates depth information injection and cross-attention control to ensure precise object placement, delivering notable improvements in spatial accuracy metrics. 
In conjunction with SmartSpatial, we present SmartSpatialEval, a comprehensive evaluation framework that bridges computational spatial accuracy with qualitative artistic assessments. 
Experimental results show that SmartSpatial significantly outperforms existing methods, setting new benchmarks for spatial fidelity in AI-driven art and creativity.
\end{abstract}

\section{Introduction}
Text-to-image generative models, particularly diffusion-based frameworks such as Stable Diffusion~\citep{rombach2021highresolution}, have achieved remarkable advances in synthesizing diverse and highly realistic images from natural language descriptions. However, despite their impressive achievements, these models frequently struggle with accurately maintaining the spatial arrangements of objects. 
This limitation becomes particularly evident when handling complex 3D spatial relationships, such as ``in front of''
and ``behind'', which require precise understanding and representation of depth and positioning. 
These inaccuracies often result in visually plausible but contextually flawed images, undermining the reliability of these models for applications demanding high spatial fidelity.

Figure~\ref{fig:teaser} shows the efficacy of SmartSpatial in enhancing 3D spatial arrangement compared to standard Stable Diffusion. 
While the left-side images, generated using Stable Diffusion, often exhibit inconsistencies in object placement and depth perception, the right-side outputs from our SmartSpatial demonstrate precise spatial alignment and structural coherence. 
By leveraging depth-aware conditioning and cross-attention refinements, SmartSpatial ensures that generated scenes adhere to the intended spatial constraints, enabling more reliable and contextually accurate text-to-image synthesis. 
This comparison underscores the necessity of spatially-aware generation techniques and highlights SmartSpatial's potential in advancing AI-driven artistic workflows.

Accurate spatial arrangement is not just a desirable feature---it is essential for critical applications like virtual scene creation, content synthesis, generating structured artistic compositions, and human-computer interaction. 
The inability of current models to consistently deliver such accuracy highlights a significant and pressing challenge in the field, underscoring the need for advanced solutions.

To bridge this gap between AI generation and artistic spatial reasoning, we propose SmartSpatial, a novel approach designed to address these limitations by incorporating 3D spatial awareness into diffusion models. 
Our method enhances object positioning precision through depth integration and cross-attention manipulation. 
By injecting 3D spatial data into ControlNet and fine-tuning cross-attention blocks, SmartSpatial achieves robust spatial arrangement capabilities guided by textual prompts. 

To comprehensively evaluate the spatial accuracy of generated images, we also propose SmartSpatialEval, an innovative evaluation framework that utilizes vision-language models (VLMs) and dependency parsing to assess spatial relationships. 
This framework provides quantitative metrics for spatial accuracy, complementing traditional image quality evaluations. 
Experimental results demonstrate that SmartSpatial significantly enhances spatial accuracy compared to existing methods, establishing a new benchmark for spatial control in text-to-image generation. Our key contributions include:
\begin{itemize}
    \item \textbf{Spatially-Aware Image Generation}: SmartSpatial integrates 3D depth information and cross-attention refinements to improve spatial precision, achieving state-of-the-art performance in both quantitative and qualitative evaluations.
    \item \textbf{Quantitative Evaluation}: SmartSpatialEval introduces robust, human-like VLM-based metrics for assessing spatial accuracy.
    \item \textbf{Dataset and Resources}: We release SpatialPrompts, a dataset designed to evaluate VLMs' 3D spatial reasoning, along with SmartSpatial and SmartSpatialEval, as research resources.\footnote{\url{https://github.com/mao-code/SmartSpatial}} 
\end{itemize}

\section{Related Works} \label{sec:relatedworks}
Recent advancements, such as MultiDiff~\citep{bartal2023multidiffusionfusingdiffusionpaths}, which employs masked noise for layout control, and eDiff-I~\citep{balaji2023ediffitexttoimagediffusionmodels}, which leverages forward guidance to improve spatial accuracy, have sought to enhance the state-of-the-art Stable Diffusion~\citep{rombach2021highresolution} framework by introducing spatial conditioning techniques. Training-free methods like Prompt-to-Prompt~\citep{hertz2022prompttopromptimageeditingcross} and pix2pix-zero~\citep{parmar2023zeroshotimagetoimagetranslation} leverage cross-attention maps for localized edits but lack holistic layout control. Extensions such as BoxDiff ~\citep{xie2023boxdifftexttoimagesynthesistrainingfree} and cross-attention backward guidance (AG) ~\citep{chen2023trainingfreelayoutcontrolcrossattention} and segmentation mask conditioning~\citep{parmar2023zeroshotimagetoimagetranslation} improve spatial precision but remain limited in complex arrangements. \citep{epstein2023diffusionselfguidancecontrollableimage} enhanced object scale and position control but struggled with fine-grained spatial accuracy.
Conditional methods improve precision by incorporating spatial guidance. ControlNet~\citep{zhang2023addingconditionalcontroltexttoimage} adds spatial conditioning through fine-tuned layers, while localized control~\citep{zhao2024localconditionalcontrollingtexttoimage} and instance-level approaches~\citep{wang2024instancediffusioninstancelevelcontrolimage} utilize bounding boxes and segmentation masks. However, these techniques often adhere rigidly to 2D layouts, limiting flexibility.

Metrics like FID~\citep{alimisis2024advancesdiffusionmodelsimage} and CLIP score~\citep{hessel2022clipscorereferencefreeevaluationmetric} prioritize visual and semantic quality but neglect spatial accuracy. 
Tools like DP-IQA~\citep{fu2024dpiqautilizingdiffusionprior} and DiffNat~\citep{roy2023diffnatimprovingdiffusionimage} focus on image quality, while the SPRIGHT dataset~\citep{chatterjee2024gettingrightimprovingspatial} highlights the need for robust spatial evaluation.  Benchmarks such as VISOR~\citep{gokhale2023benchmarkingspatialrelationshipstexttoimage} and its evaluation framework primarily address two-dimensional spatial accuracy, leaving a significant gap in evaluating more complex, three-dimensional spatial relationships. These works highlights the critical limitation of current tools in assessing spatial arrangements effectively, particularly in scenarios requiring robust 3D spatial understanding.

Our work advances 3D spatial arrangement through cross-attention manipulation and 3D conditioning, surpassing limitations of planar-focused methods like \citep{chen2023trainingfreelayoutcontrolcrossattention} and rigid controls in ControlNet~\citep{zhang2023addingconditionalcontroltexttoimage}. 
We further address the gap in evaluation by introducing SmartSpatialEval, a comprehensive tool for assessing spatial accuracy in generated images.

\section{SmartSpatial}
\begin{figure}[!h]
    \centering
    \includegraphics[width=\linewidth]{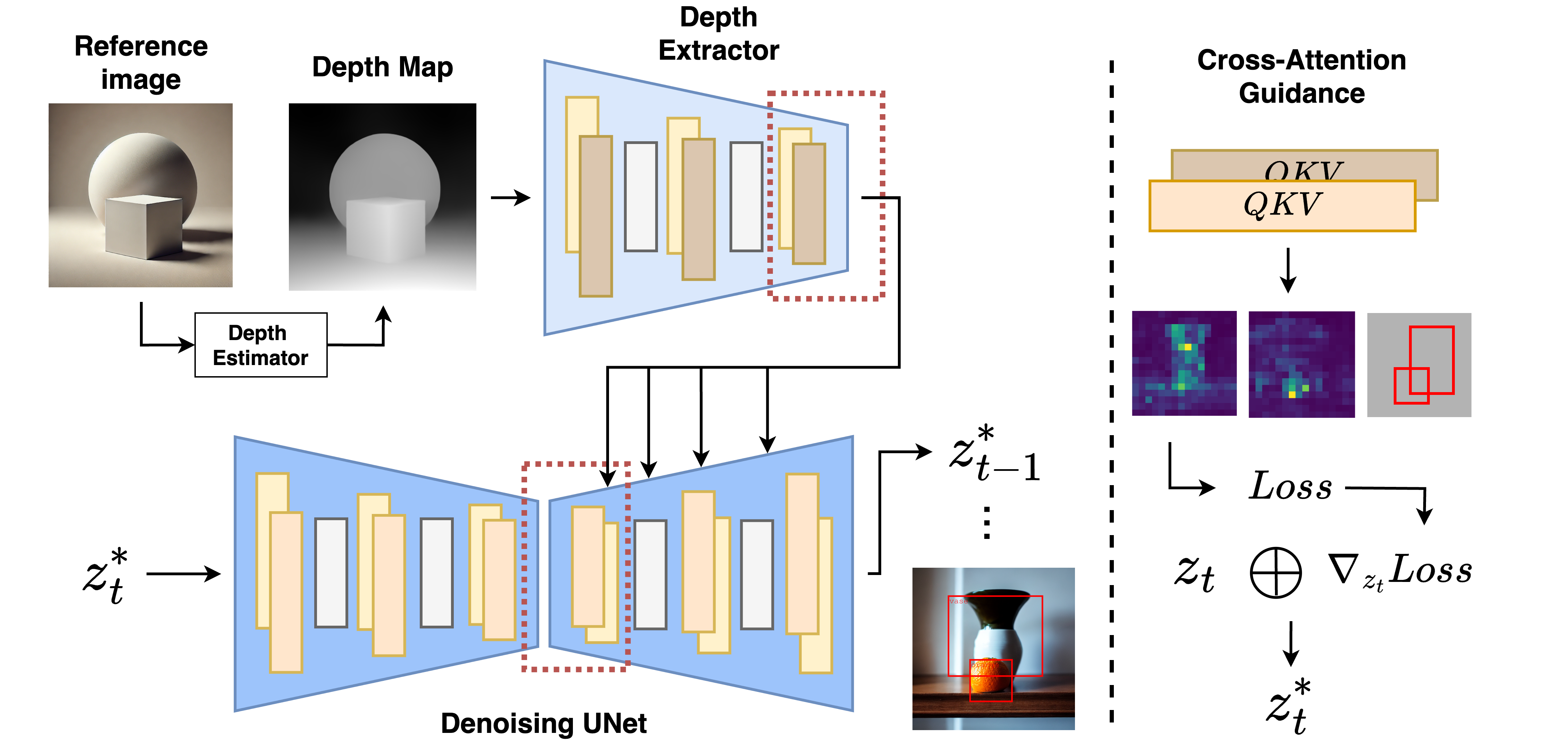}
    \caption{The SmartSpatial process involves depth extraction and cross-attention guidance. 
    A reference image (``The ball is behind the box'') generates a depth map via a depth estimator, injected into the denoising UNet by the depth extractor. 
    Cross-attention blocks from both the depth extractor and denoising UNet are extracted to guide object focus, ultimately generating an image of ``A vase is behind an orange.''}
    \label{fig:smart_spatial}
\end{figure}

SmartSpatial is a 3D-aware enhancement for Stable Diffusion models. 
As illustrated in Figure~\ref{fig:smart_spatial}, we propose a novel approach that integrates 3D spatial data into ControlNet with attention-guided mechanisms, enabling precise spatial arrangement while maintaining high image quality. 

\subsection{3D Information Integration and Attention-Guided Control} \label{sec:method_smart_spatial}
To enhance 3D spatial arrangement in Stable Diffusion, we integrate depth-aware conditioning and attention-guided control. 
Depth information is injected via ControlNet, enriching spatial representation, while refined cross-attention mechanisms improve object placement. 
A tailored loss function optimizes spatial coherence, ensuring alignment with textual prompts. 
The details are given in Section~\ref{sec:method_depth}, 
Section~\ref{sec:method_attention}, and Section~\ref{sec:method_loss}, respectively. 

\subsection{Depth Information Injection} \label{sec:method_depth}
To capture 3D spatial relationships such as ``in front of'' and ``behind'', we select a reference image and employ a depth estimator to generate a corresponding depth map. 
Note that the reference image can be any image where the objects represent a spatial relationship, making it adaptable to various scenarios. 
It is not confined to a specific image but serves as a general guiding example. 
For instance, the reference image in Figure~\ref{fig:smart_spatial}, depicting ``A ball is behind a box,'' can be applied broadly to cases involving the ``behind'' relationship. 

Reference images can be automatically created using a 3D drawing toolkit such as Matplotlib or Blender. 
These toolkits are particularly well-suited for generating simple 3D scenes, as objects like ``ball'' and ``box'' are relatively easy to model in such environments. 
This makes them a promising and accessible source for creating reference images that can then be converted into depth maps.

The generated depth map is subsequently processed by a depth extractor, utilizing ControlNet~\citep{zhang2023addingconditionalcontroltexttoimage}, to extract depth features. 
The extracted depth information is subsequently integrated into the upsampling blocks of the denoising UNet, enriching the model with precise spatial data.

\subsection{Attention Block Selection}\label{sec:method_attention} 
ControlNet often rigidly constrains generated images to the reference input, so we mitigate this by modifying the cross-attention blocks. 
Specifically, we select the mid-cross-attention block in the depth extractor along with the mid and first up-sampling cross-attention blocks in the denoising UNet. 
This configuration has been shown to provide optimal performance, as demonstrated in \citet{chen2023trainingfreelayoutcontrolcrossattention}, enhancing the model's ability to guide spatial awareness and object placement.
In Figure~\ref{fig:smart_spatial}, for example, the model is guided to identify the ``ball'' as the vase and the ``box'' as the orange.

\subsection{Loss Function and Attention Guidance}\label{sec:method_loss}  
Our objective is to fine-tune the latent space to ensure high attention weights within designated regions. 
Inspired by \citet{chen2023trainingfreelayoutcontrolcrossattention}, we extract attention maps \(A_i\) for the \(i\)-th token from the depth extractor and the denoising UNet. 
To confine \(A_i\) predominantly within the specified bounding box \(b_i\), we adopt the following loss function:
\begin{equation}
    L = \sum_{b_i \in B}\left(1 - \frac{\sum_{p \in b_i} A_{p,i}}{\sum_{p} A_{p,i}}\right)^2
\end{equation} Here, \(A_{p,i}\) denotes the attention values at pixel \(p\) for token \(i\), and \(B\) is the set of all bounding boxes.


\begin{equation}
\label{eq:loss}
\begin{aligned}
    v_i^{(t)} &\leftarrow m\,v_{i-1}^{(t)} \;-\; \eta\,\nabla_{z_{i}^{(t)}} L\\
    z_{i+1}^{(t)} &\leftarrow z_{i}^{(t)} \;+\; v_i^{(t)}
\end{aligned}
\end{equation}
where \( m \) is the momentum coefficient, \( \eta \) is the learning rate, \( t \) denotes the current denoising step, and \( i \) is the iteration index for cross-attention guidance. The variable \( z_i^{(t)} \) represents the attention-guided latent variable at iteration \( i \) of denoising step \( t \). The iteration index \( i \) ranges from \( 1 \) to \( K \), where \( K \) is the maximum number of iterations. \( K \) can be predefined or dynamically determined by stopping when the total loss falls below a set threshold. Therefore, the final attention guided latent at denoising step \( t \) will be \( z_K^{(t)} \)

Additionally, we incorporate a ControlNet specific term in the overall loss function to ensure coherent guidance across the entire model:
\begin{equation}
    L_{\text{total}} = \alpha L_{\text{unet}} + \beta L_{\text{control}}
\end{equation}
Here, \( L_{\text{unet}} \) and \( L_{\text{control}} \) represent the loss components for the UNet and ControlNet, respectively. The coefficients \( \alpha \) and \( \beta \) are weighting factors that balance the contributions of each term. The calculations for \( L_{\text{unet}} \) and \( L_{\text{control}} \) are consistent with those in Eq.~\ref{eq:loss}, with the distinction that the cross-attention maps are extracted from different models—UNet and ControlNet, respectively.

\section{SmartSpatialEval}
SmartSpatialEval is a novel framework that leverages VLMs, dependency parsing, and graph-based spatial representations to quantitatively assess spatial relationships against ground truth data. 
It provides a structured and objective evaluation of 3D spatial fidelity, addressing a critical gap in text-to-image generative models.

\begin{figure}[bth]
    \centering
    \includegraphics[width=\linewidth]{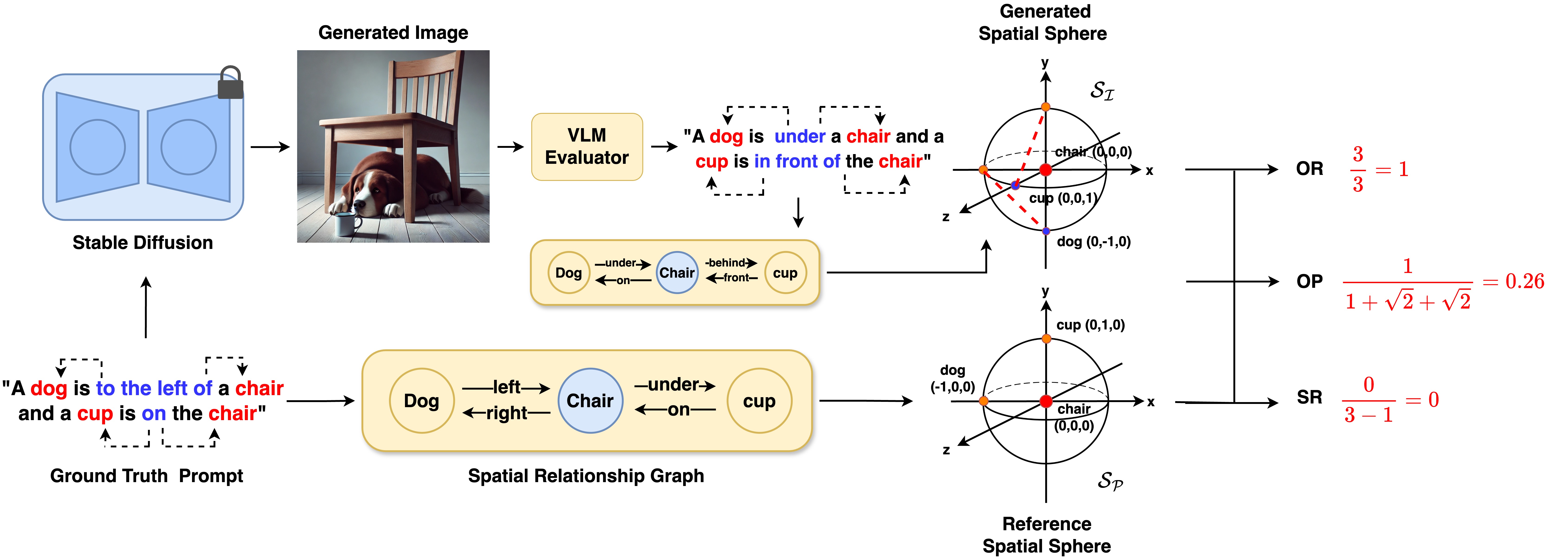}
    \caption{SmartSpatialEval evaluates spatial accuracy for the image generated from the prompt ``A dog is to the left of a chair, and a cup is on the chair'' by comparing the spatial sphere \( \mathcal{S_I} \), constructed from the spatial relationship graph of the generated image, with the reference spatial sphere \( \mathcal{S_P} \), derived from the spatial relationship graph of the original textual prompt.}
    \label{fig:smart_spatial_eval}
\end{figure}

As illustrated in Figure~\ref{fig:smart_spatial_eval}, SmartSpatialEval evaluates an image generated for the prompt ``A dog is to the left of a chair, and a cup is on the chair'' by constructing a spatial sphere $\mathcal{S_I}$ from a graph that encodes the spatial relationships among objects in the image. 
This is then compared to a reference spatial sphere $\mathcal{S_P}$, derived from the graph representing the spatial relationships in the original textual prompt, enabling precise spatial alignment assessment.

\begin{figure}[tbh]
    \centering
    \includegraphics[width=0.6\linewidth]{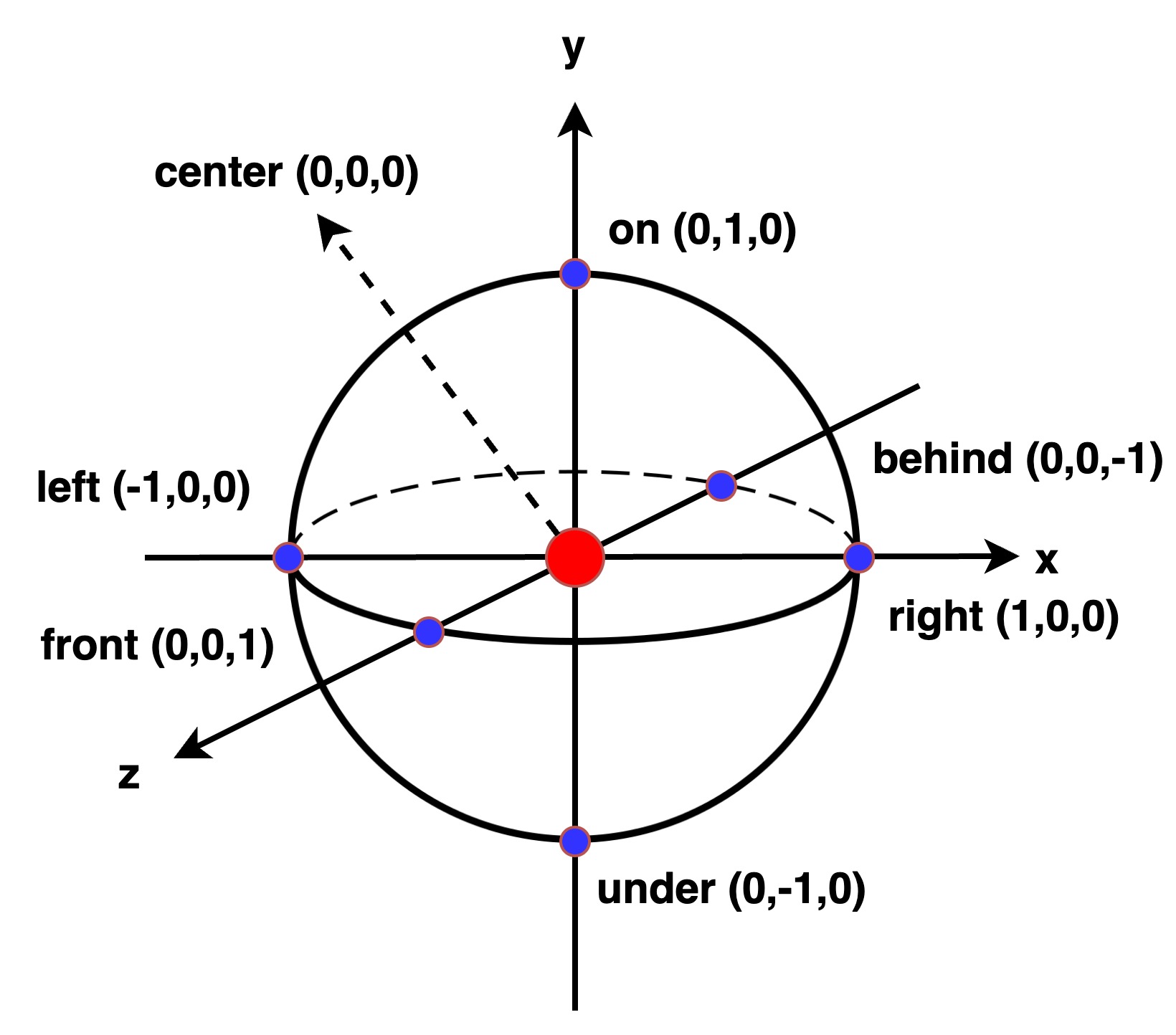}
    \caption{The Spatial Sphere model quantifies positional language, representing each point's relative relationship to the center.}
    \label{fig:spatial_sphere}
\end{figure}

\subsection{Framework and Metrics} 
\label{sec:method_smart_spatial_eval}

To represent spatial relationships among objects in a generated image \(\mathcal{I}\) corresponding to a textual prompt \(\mathcal{P}\), we first employ a VLM (ChatGPT-4o) to generate a textual description of the spatial relationships in \(\mathcal{I}\). 
This description is then parsed into a spatial relationship graph \( \mathcal{G_I} \) using dependency parsing~\citep{honnibal2020spacy}, capturing the spatial structure among objects. 
Next, we transform \( \mathcal{G_I} \) into a spatial sphere \( \mathcal{S_I} \), as illustrated in Figure~\ref{fig:spatial_sphere}.
In this representation, the center object is positioned at the core of the sphere, with other objects arranged based on their relative spatial relationships, enabling structured comparisons.  

Similarly, we construct the reference spatial sphere \( \mathcal{S_P} \) by first extracting a spatial relationship graph \( \mathcal{G_P} \) from the original textual prompt \( \mathcal{P} \) using dependency parsing, then transforming it into \( \mathcal{S_P} \).  
To compare \( \mathcal{S_I} \) and \( \mathcal{S_P} \), we designate the center object, that is identified as the root of the dependency parse tree of $\mathcal{P}$, and use breadth-first search (BFS) to extract the shortest paths from this center to all other objects. 
We then compute three key spatial accuracy scores:  

\begin{enumerate}
    \item \textbf{Object Recognition (OR) Score} measures the model's ability to generate all objects specified in the prompt. Image generation models often fail to include all objects, necessitating this metric:  
\begin{equation}
    OR = \frac{N_{\mathcal{I}}}{N_{\mathcal{P}}}
\end{equation}\label{sec:method_obj_score}  

where \( N_{\mathcal{I}} \) is the count of correctly identified objects in $\mathcal{I}$, and \( N_{\mathcal{P}} \) is the total number of objects specified in the prompt $\mathcal{P}$.  

\item \textbf{Object Proximity (OP) Score} measures how accurately the generated objects are positioned relative to their expected locations by computing the inverse of the total Euclidean distances: 
\begin{equation}
    OP = \frac{1}{1 + \sum_{i=1}^{N_{\mathcal{P}}} \|\mathbf{r}_i - \mathbf{o}_i\|_2}
\end{equation} \label{sec:method_d_score}  

where \( \mathbf{r}_i \) represents the reference 3D position of object \( i \) as specified in the prompt \( \mathcal{P} \), and \( \mathbf{o}_i \) represents the generated 3D position of object \( i \) in the generated image \( \mathcal{I} \).  

If an object is missing in \( \mathcal{I} \), its generated position \( \mathbf{o}_i \) is assigned to a distant outlier location, ensuring that the corresponding Euclidean distance \( \|\mathbf{r}_i - \mathbf{o}_i\|_2 \) remains large. 
This penalizes missing objects, leading to a lower OP score, effectively capturing the model's ability to generate objects at the correct spatial locations.  

\item \textbf{Spatial Relationship (SR) Score} measures how accurately the generated image $\mathcal{I}$ preserves the spatial relationships specified in the prompt $\mathcal{P}$. 
This score evaluates relative positioning based on a spatial representation.

\begin{equation}
    SR = \frac{M_{\mathcal{I}}}{N_{\mathcal{P}} - 1}
\end{equation}
\label{sec:method_sr_score}  

where \( M_{\mathcal{I}} \) represents the number of correctly identified spatial relationships in the generated image $\mathcal{I}$, and \(N_{\mathcal{P}} \) is the total number of objects specified in the prompt $\mathcal{P}$.  
Since spatial relationships are defined as the relationships between the center object and each of the remaining $N_{\mathcal{P}} - 1$ objects, the denominator reflects the expected number of valid spatial relationships in the reference spatial sphere $\mathcal{S_P}$.
\end{enumerate}

For all the three metrics, a score of 1.0 indicates perfect adherence to the specified spatial constraints, while lower values suggest deviations from the intended spatial arrangement.
In Figure~\ref{fig:smart_spatial_eval}, all three objects (dog, chair, and cup) are successfully rendered in \( \mathcal{I} \), yielding an Object Recognition (OR) Score of 1.  
The chair, identified as the center object by the dependency parser, is positioned at the origin \( (0, 0, 0) \). 
However, the spatial relationships of the other objects deviate from the expected configuration:  
\begin{itemize}
    \item The dog is misplaced \textit{under} the chair at \( (0, -1, 0) \) instead of the expected position \textit{to the left} at \( (-1, 0, 0) \), resulting a distance of ${\| (0, -1, 0) - (-1, 0, 0) \|_2 = \sqrt{2}}$.  
    \item The cup is misplaced \textit{in front of} the chair at \( (0, 0, 1) \) rather than \textit{on top of} it at \( (0, 1, 0) \), resulting a distance of ${\| (0, 1, 0) - (0, 0, 1) \|_2 = \sqrt{2}}$.
\end{itemize}
As a result, the Object Proximity (OP) Score is $\frac{1}{1 + \sqrt{2} + \sqrt{2}} = 0.2612$.
Since the two expected spatial relationships in \( \mathcal{P} \) (\texttt{left(dog, chair)} and \texttt{on(cup, chair)}) are both incorrectly generated in \( \mathcal{I} \), the Spatial Relationship (SR) Score is 0.  

Unlike existing metrics such as CLIP, IoU, and mAP, which primarily assess image quality or layout precision, SmartSpatialEval specifically evaluates 3D spatial arrangements. 
Our Proximity Score and Spatial Relationship Score leverage VLM-based observations to simulate human perception, assessing images in terms of complex 3D spatial relationships (e.g., front, behind, left, right, above, below). 
This approach ensures a more precise and contextually meaningful evaluation of spatial consistency in text-to-image generation.  

Moreover, SmartSpatialEval can also serve as a reinforcement learning reward signal, enabling reinforcement learning methods like DDPO~\citep{black2024trainingdiffusionmodelsreinforcement} to optimize diffusion models for spatial reasoning. This expands its role from benchmarking to training AI for diverse spatial tasks.

\begin{table*}[!htb]
\centering
\begin{adjustbox}{max width=\linewidth}
\begin{tabular}{llccccccc}
\toprule
\textbf{Dataset} & \textbf{Method} & \textbf{CLIP $\uparrow$} & \textbf{mAP@0.5 $\uparrow$} & \textbf{IoU $\uparrow$} & \textbf{OP $\uparrow$} & \textbf{SR $\uparrow$} & \textbf{OR $\uparrow$} & \textbf{OP+OR $\uparrow$} \\
\midrule
\multirow{7}{*}{SpatialPrompts} 
& MultiDiff       & 0.236 & 0.006 & 0.020 & 0.034 & 0.017 & 0.229 & 0.132 \\
& eDiff-I         & \textbf{0.311} & 0.010 & 0.019 & 0.338 & 0.208 & \textbf{0.796} & 0.567 \\
& BoxDiff         & 0.308 & 0.041 & 0.075 & 0.287 & 0.183 & 0.725 & 0.506 \\
& SD              & 0.295 & 0.019 & 0.039 & 0.249 & 0.167 & 0.700 & 0.475 \\
& SD+AG           & 0.305 & 0.132 & 0.223 & 0.380 & 0.300 & 0.746 & 0.563 \\
& SD+ControlNet   & 0.296 & 0.051 & 0.099 & 0.200 & 0.108 & 0.683 & 0.442 \\
& SmartSpatial (Ours) & 0.303 & \textbf{0.311} & \textbf{0.434} & \textbf{0.433} & \textbf{0.358} & 0.775 & \textbf{0.604} \\
\midrule
\multirow{7}{*}{COCO2017} 
& MultiDiff       & 0.171 & 0.000 & 0.001 & 0.000 & 0.000 & 0.030 & 0.015 \\
& eDiff-I         & \textbf{0.325} & 0.022 & 0.034 & 0.166 & 0.073 & 0.654 & 0.410 \\
& BoxDiff         & 0.317 & 0.029 & 0.043 & 0.108 & 0.043 & 0.594 & 0.351 \\
& SD              & 0.314 & 0.013 & 0.024 & 0.084 & 0.026 & 0.567 & 0.325 \\
& SD+AG           & 0.321 & 0.087 & 0.130 & 0.227 & 0.153 & 0.660 & 0.443 \\
& SD+ControlNet   & 0.314 & 0.028 & 0.048 & 0.083 & 0.028 & 0.566 & 0.324 \\
& SmartSpatial (Ours) & 0.312 & \textbf{0.207} & \textbf{0.309} & \textbf{0.286} & \textbf{0.218} & \textbf{0.672} & \textbf{0.479} \\
\midrule
\multirow{7}{*}{VISOR} 
& MultiDiff       & 0.241 & 0.000 & 0.000 & 0.000 & 0.000 & 0.020 & 0.010 \\
& eDiff-I         & \textbf{0.325} & 0.023 & 0.038 & 0.154 & 0.063 & 0.656 & 0.405 \\
& BoxDiff         & 0.320 & 0.022 & 0.036 & 0.111 & 0.042 & 0.606 & 0.358 \\
& SD              & 0.315 & 0.011 & 0.017 & 0.088 & 0.022 & 0.574 & 0.332 \\
& SD+AG           & 0.326 & 0.103 & 0.150 & 0.279 & 0.213 & 0.688 & 0.483 \\
& SD+ControlNet   & 0.316 & 0.027 & 0.046 & 0.088 & 0.029 & 0.569 & 0.328 \\
& SmartSpatial (Ours) & 0.312 & \textbf{0.219} & \textbf{0.324} & \textbf{0.352} & \textbf{0.302} & \textbf{0.700} & \textbf{0.526} \\
\bottomrule
\end{tabular}
\end{adjustbox}
\caption{Experimental results on SpatialPrompts, COCO2017 and VISOR datasets}
\label{tab:performance}
\end{table*}

\begin{figure*}[!ht]
    \centering
    \includegraphics[width=\linewidth]{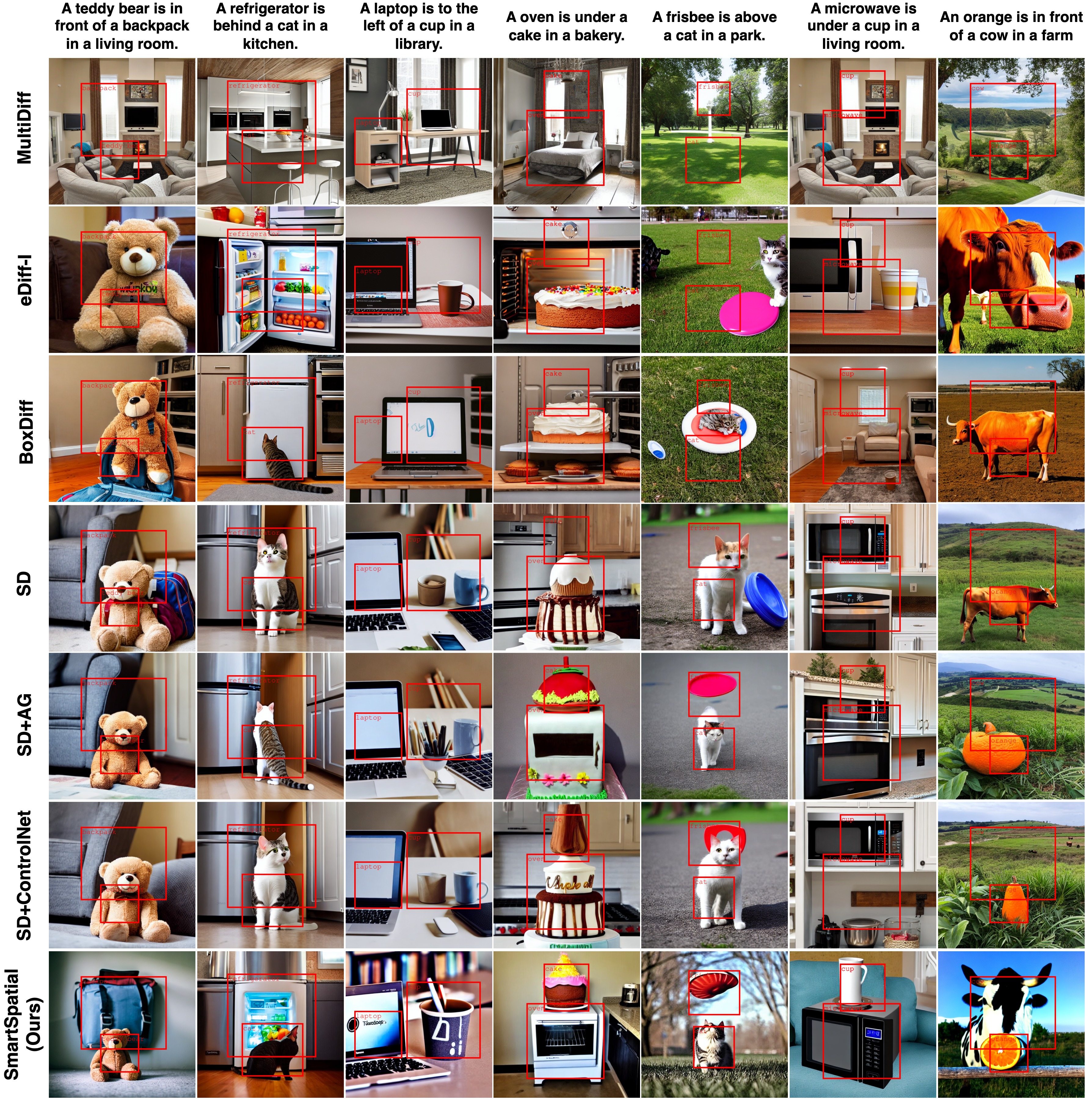}
    \caption{Qualitative comparison of spatial control methods. All generated images are based on 
    SpatialPrompts with bounding boxes derived from reference images.
    Our approach exhibits superior spatial control compared to other guidance methods.}
    \label{fig:qualitive}
\end{figure*}

\section{Experiments}
\subsection{Experimental Setup}\label{sec:exp_ss}
We evaluate our SmartSpatial on three datasets as follows. 

\begin{itemize}
    \item \textbf{SpatialPrompts} is a custom dataset comprising 120 hand-crafted prompts designed to test SmartSpatial's spatial reasoning abilities. 
    These prompts represent realistic and commonly occurring scene scenarios, such as ``A bicycle is in front of a car at a traffic signal.''
    SpatialPrompts covers eight spatial positions (i.e., front, behind, left, right, on, under, above, below) in the 3D setting, with 15 examples for each category. 
    \item 1,000 samples derived from \textbf{COCO2017}~\citep{lin2015microsoftcococommonobjects}. 
    To make this dataset more sutible for our scenario, we sampled two unique objects from each of the 80 categories in COCO2017, paired them with one spatial term from the eight spatial positions defined in SpatialPrompts, and combined them with a background term selected from a custom set of 10 types (e.g., \textit{park, library}). 
    This process resulted in a total of $80 \times 8 \times 10 = 6,400$ combinations, from which we randomly selected 1,000 instances.
    The random sampling introduces more complex, uncommon, and surreal prompts, making this dataset particularly challenging for spatially controlled image generation tasks.
    \item 1,000 samples derived from \textbf{VISOR}~\citep{gokhale2023benchmarkingspatialrelationshipstexttoimage}. 
    Since VISOR contains only two-dimensional spatial relationships, we randomly selected 336 instances and replaced their spatial terms with three-dimensional spatial descriptors, such as \textit{front} and \textit{behind}. 
    This adjustment enriches the dataset by introducing additional complexity and testing the ability of SmartSpatial to handle three-dimensional spatial reasoning tasks effectively.
\end{itemize}

We compare SmartSpatial with several state-of-the-art models, including MultiDiff~\citep{bartal2023multidiffusionfusingdiffusionpaths}, eDiff-I~\citep{zhang2023talefeaturesstablediffusion}, BoxDiff~\citep{xie2023boxdifftexttoimagesynthesistrainingfree}, SD~\citep{rombach2021highresolution}, SD+AG~\citep{chen2023trainingfreelayoutcontrolcrossattention}, and SD+ControlNet~\citep{zhang2023addingconditionalcontroltexttoimage}. 
These baseline models provide a diverse set of approaches for text-to-image synthesis, ranging from diffusion-based methods to techniques incorporating additional conditional controls, allowing for comprehensive and robust comparisons.

In addition to the three metrics provided by SmartSpatialEval, we also adopt three widely-used metrics in the experiments, including CLIPScore~\citep{hessel2022clipscorereferencefreeevaluationmetric} for image-text alignment, IoU ~\citep{redmon2016lookonceunifiedrealtime}, and mAP@0.5 for object layour control accuracy. 

All experiments were conducted using Stable Diffusion v1.5~\citep{rombach2021highresolution} as the backbone model for text-to-image generation.
The experiments were executed on a single Tesla V100-SXM2 GPU with 32GB memory. 
We set the random seed to 42 and employed a cross-attention guidance loss threshold of 0.5. 

\subsection{Results}
As summarized in Table~\ref{tab:performance}, our proposed approach, SmartSpatial, consistently demonstrates superior performance across most metrics and datasets.
Notably, the improvements in IoU and mAP metrics underscore the enhanced capability of SmartSpatial in layout control. Additionally, higher OP, SR, and OR scores highlight its advanced understanding of spatial knowledge, enabling the generation of images with accurate object presence and spatial arrangements. 
While a minor decrease in CLIPScore indicates a slight trade-off between precise spatial control and overall image quality, the competitive CLIPScore achieved suggests that SmartSpatial maintains acceptable visual quality.
Statistical significance tests confirmed that these  performance differences are not significant across all datasets ($p > 0.05$).

Furthermore, on more complex and diverse datasets such as COCO2017 and VISOR, where overall scores are lower due to intricate scenes and object relationships, SmartSpatial continues to exhibit robust spatial control capabilities. 
This highlights its adaptability and effectiveness, even in challenging scenarios involving complex spatial dynamics.

The qualitative results presented in Figure~\ref{fig:qualitive} further validate our quantitative findings. 
The baseline Stable Diffusion model often struggles with issues such as missing objects and inaccurate spatial relationships. 
Similarly, other layout control methods occasionally fail to maintain spatial coherence, particularly in complex scenarios involving 3D relationships (e.g., objects positioned ``in front of others'' or ``behind others'') or unconventional arrangements, such as ``an orange in front of a cow on a farm.''
In contrast, SmartSpatial consistently preserves spatial relationships, effectively managing spatial prompts across a wide variety of conditions. 
This highlights its reliability and practical applicability in handling both common and uncommon spatial arrangements with remarkable precision.

\subsection{Ablation Study}
To evaluate the effectiveness of each component in our proposed system, we conducted an ablation study on the VISOR dataset, comprising 1,000 instances. 
Our system is decomposed into three key components:  Cross-Attention Guidance (AG), ControlNet (CN), and Cross-Attention Guidance with ControlNet (CNAG). 

\begin{figure}[!bth]
    \centering
    \includegraphics[width=\linewidth]{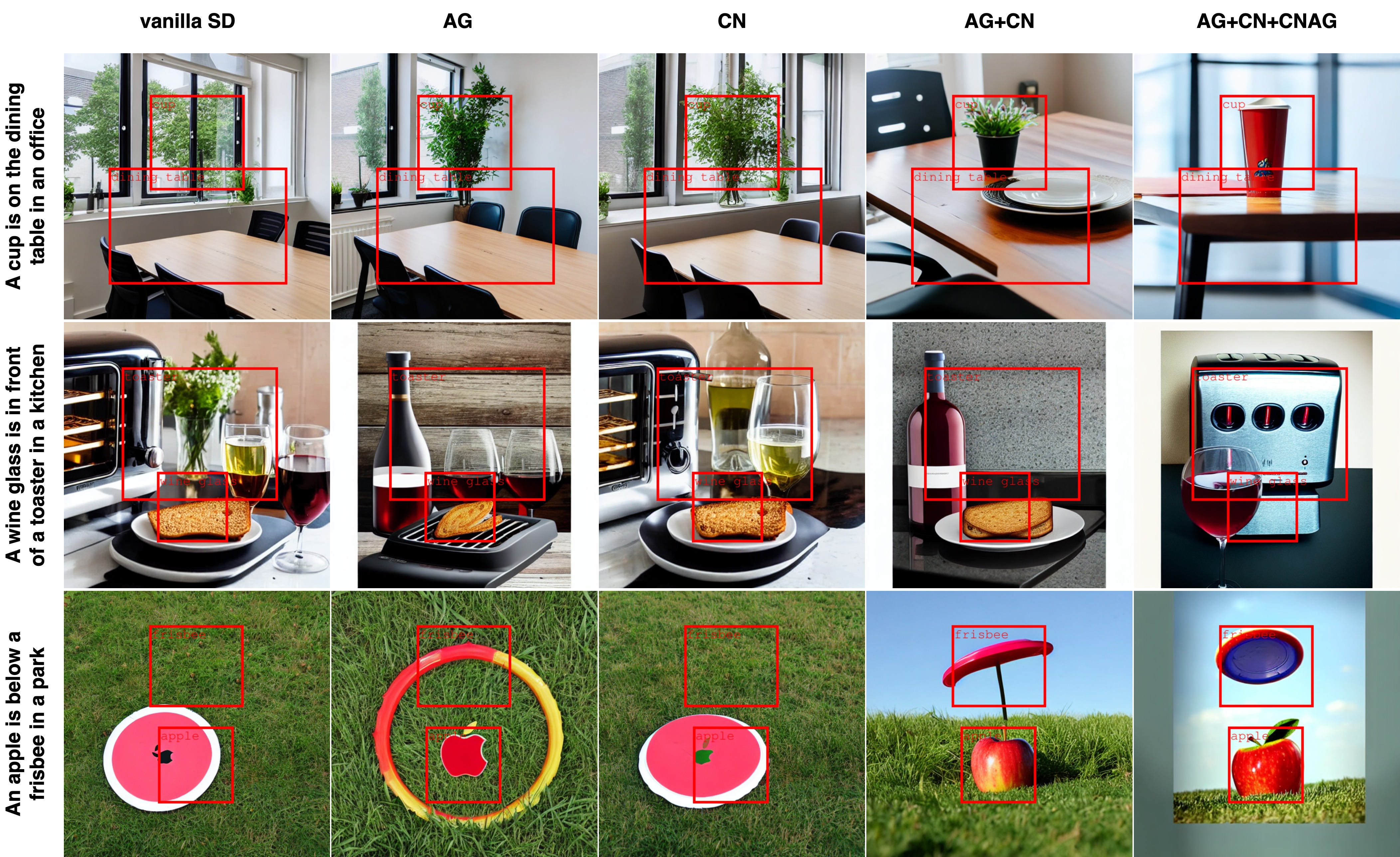}
    \caption{Results for different configurations. Incorporating all components (AG + CN + CNAG) achieves the best spatial control.}
    \label{fig:ablation_qualitve}
\end{figure}

Table~\ref{tab:ablation_study} highlights that the best spatial control results are achieved when all three components (AG, CN, and CNAG) are employed. 
Although a slight decrease in the CLIP score is observed in the AG + CN + CNAG configuration, the difference is not statistically significant ($p > 0.05$). 
The qualitive results are shown in Figure~\ref{fig:ablation_qualitve}. 
Both the quantitive and qualitive results demonstrate that our system enhances spatial awareness and significantly improves layout control for Stable Diffusion while maintaining competitive image quality.

\begin{table}[!bth]
\centering
\begin{adjustbox}{max width=\linewidth}
\begin{tabular}{ccccccccc}
\toprule
\textbf{AG} & \textbf{CN} & \textbf{CNAG} & \textbf{CLIP} & \textbf{mAP} & \textbf{IoU} & \textbf{OP} & \textbf{SR} & \textbf{OR} \\
\midrule
-- & -- & -- & 0.315 & 0.011 & 0.017 & 0.088 & 0.022 & 0.575 \\
\checkmark & -- & -- & \textbf{0.326} & 0.103 & 0.150 & 0.279 & 0.213 & 0.688 \\
-- & \checkmark & -- & 0.316 & 0.027 & 0.046 & 0.088 & 0.029 & 0.569 \\
\checkmark & \checkmark & -- & 0.325 & 0.145 & 0.206 & 0.269 & 0.210 & 0.668 \\
\checkmark & \checkmark & \checkmark & 0.312 & \textbf{0.219} & \textbf{0.324} & \textbf{0.352} & \textbf{0.302} & \textbf{0.700} \\
\bottomrule
\end{tabular}
\end{adjustbox}
\caption{Results of ablation analysis}
\label{tab:ablation_study}
\end{table}

\section{Applications in Art and Design} 
The proposed SmartSpatial method opens up new possibilities in AI-driven art and design. 
For instance, in marketing and advertising, precise spatial arrangements of objects (e.g., products on a table or models interacting with items) are often critical for creating impactful visuals that align with strategic goals. 
SmartSpatial excels in these scenarios by enabling accurate and flexible spatial control.

\begin{figure}[!tbhp]
    \centering
    \includegraphics[width=0.9\linewidth]{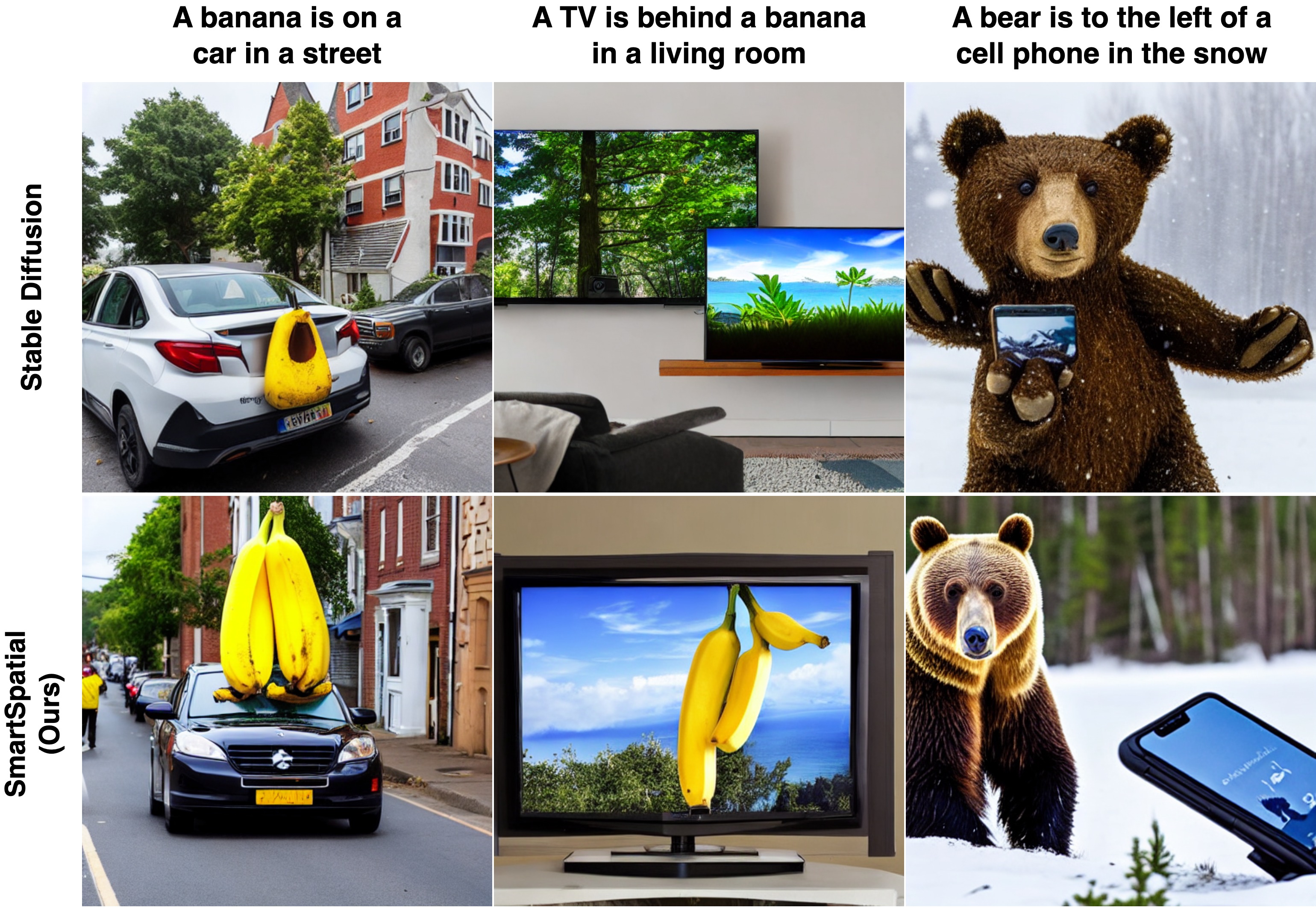}
    \caption{Surreal scene generation with precise spatial control. Traditional Stable Diffusion often fails to include all objects from the prompt or arranges them incorrectly. 
    In contrast, SmartSpatial accurately places even unrelated objects in the specified configuration, preserving both composition and spatial coherence.}
    \label{fig:surreal}
\end{figure}

In the realm of surreal art, where unconventional or unrelated objects are often juxtaposed, SmartSpatial provides artists with a powerful tool for generating visually striking compositions with precise spatial arrangements. 
As demonstrated in Figure~\ref{fig:surreal}, SmartSpatial supports the creation of unique and imaginative scenes, enabling both artists and marketing professionals to craft compelling visual content that pushes creative boundaries. By offering robust 3D spatial control and surreal image generation capabilities, SmartSpatial represents a valuable contribution to the fields of AI-assisted art and design.

Additionally, SmartSpatial can generate visual-text spatial pair datasets to enhance the spatial intelligence and inference ability of VLMs. 
The lack of such datasets limits VLMs' ability to understand spatial relationships, but by leveraging SmartSpatial's 3D-aware conditioning, we can systematically create high-quality spatial datasets to fill this gap. 
This aligns with dataset like Synergistic-General-Multimodal Pairs~\citep{huang2024synergistic}, which showed that integrating text-to-image models with VLMs improves multimodal learning. 
A SmartSpatial-generated dataset can similarly enhance spatial reasoning in VLMs, benefiting AI-driven art and design.
Moreover, by representing and quantifying spatial coherence and evaluating 3D spatial consistency, SmartSpatialEval facilitates structured assessments of AI-generated artistic compositions

\section{Conclusions}
This work introduced SmartSpatial, a novel approach to enhance 3D spatial arrangement in text-to-image generative models, and SmartSpatialEval, an innovative framework for evaluating spatial accuracy. 
By integrating 3D spatial information and refining cross-attention mechanisms, SmartSpatial improves spatial precision while maintaining image quality. 
Our contributions pave the way for more reliable and context-aware image synthesis in applications requiring high spatial fidelity.

\section*{Ethics Statement}
This work does not present any ethical concerns. All datasets used in this study are either publicly available or automatically generated, ensuring compliance with ethical and legal standards. Additionally, this research does not involve human subjects, personal data, or any sensitive information. No human annotations or interventions were required beyond standard benchmarking practices.

\section*{Acknowledgments}
This work was partially supported by National Science and Technology Council (NSTC), Taiwan, under the grant 112-2221-E-001-016-MY3, by Academia Sinica, under the grant 236d-1120205, and by National Center for High-performance Computing (NCHC), National Applied Research Laboratories (NARLabs), and NSTC under the project ``Trustworthy AI Dialog Engine (TAIDE).'' 

\bibliographystyle{named}
\bibliography{smartspatial}

\end{document}